\documentclass[10pt]{article} 
\usepackage[preprint]{tmlr}

\usepackage{microtype}
\usepackage{graphicx}
\usepackage{subfigure}
\usepackage{booktabs} 
\usepackage{siunitx}   
\usepackage{tabularx}  
\usepackage{multirow}  
\usepackage{paralist}

\usepackage{enumitem}
\usepackage{paralist}
\usepackage{makecell}

\usepackage{etoolbox}  
\usepackage{threeparttable}
\usepackage{listings}
\usepackage[most]{tcolorbox}
\usepackage{xcolor}

\newtcolorbox{promptbox}{
    enhanced,
    boxrule=0.8pt,    
    colframe=black!80,    
    colback=gray!5,       
    arc=10pt,           
    left=15pt,          
    right=15pt,            
    top=10pt,           
    bottom=10pt,         
    fontupper=\small\rmfamily,
}
\robustify\bfseries
\usepackage{hyperref}

\usepackage{amsmath}
\usepackage{amssymb}
\usepackage{mathtools}
\usepackage{amsthm}

\usepackage[capitalize,noabbrev]{cleveref}

\theoremstyle{plain}

\theoremstyle{definition}

\theoremstyle{remark}

\usepackage[textsize=tiny]{todonotes}

\usepackage{algorithm}
\usepackage{wrapfig,lipsum}

\usepackage[utf8]{inputenc} 
\usepackage[T1]{fontenc}    
\usepackage{hyperref}       
\usepackage{url}            
\usepackage{booktabs}       
\usepackage{amsfonts}       
\usepackage{nicefrac}       
\usepackage{microtype}      
\usepackage{xcolor}         

\usepackage{rotating}
\usepackage{tcolorbox}
\usepackage{algorithm}
\usepackage{algpseudocode}
\usepackage{amsmath}
\usepackage{algorithmicx}
\usepackage{booktabs}
\usepackage{multirow}
\usepackage{multicol}
\usepackage{amsfonts}
\usepackage{cleveref}

\usepackage{wrapfig,lipsum}

\usepackage{caption}
\usepackage{subcaption}

\usepackage[textsize=tiny]{todonotes}

\title{Multimodal Thinking with Renderable Programs}

\author{%
  \name Sunli Chen \\
  \addr University of Massachusetts Amherst
  \AND
  \name Ding Zhong \\
  \name Ziqiao Ma \\
  \addr University of Michigan
  \AND
  \name Jiaxin Liu \\
  \addr University of Illinois Urbana-Champaign
  \AND
  \name Zeyuan Yang \\
  \name Hao Zhang \\
  \addr University of Massachusetts Amherst
  \AND
  \name Lie Lu \\
  \addr Dolby Laboratories
  \AND
  \name Joyce Chai \\
  \addr University of Michigan
  \AND
  \name Chuang Gan \\
  \addr University of Massachusetts Amherst
}

\def\month{MM}  
\def\year{YYYY} 
\def\openreview{\url{https://openreview.net/forum?id=XXXX}} 

\begin{document}

\maketitle


\begin{abstract}

Current vision-language models (VLMs) excel at visual content understanding and text-based reasoning, yet their structure limits the advancement of incorporating images into the reasoning chain. 
Though Omnimodal models have made efforts in unifying text and image generation, they focus on visual tasks in the open-domain, lacking tractability due to rasterized or latent representations of images. 
We introduce SVGLM, a framework that uses scalable vector graphics (SVG) primitives to connect text and image in reasoning tasks. 
We exploit the duality of SVG as both image description and text instructions, yielding a more compact, interpretable solution to equip general VLMs with the capability of generating images within the reasoning process. We provide a large curated dataset of SVG-based image editing dataset, as well as the paradigm to tune open-source VLMs. Experiments on a mathematical reasoning benchmark demonstrate that SVGLM achieves strong SVG generation power as well as think-with-image intelligence. Our results highlight SVG as a suitable medium for building more robust digital domain agents, bridging the gap between text-based thinking and pixel-based images.

\end{abstract}

\begin{figure*}[!h]
  \centering
  \includegraphics[width=\textwidth]{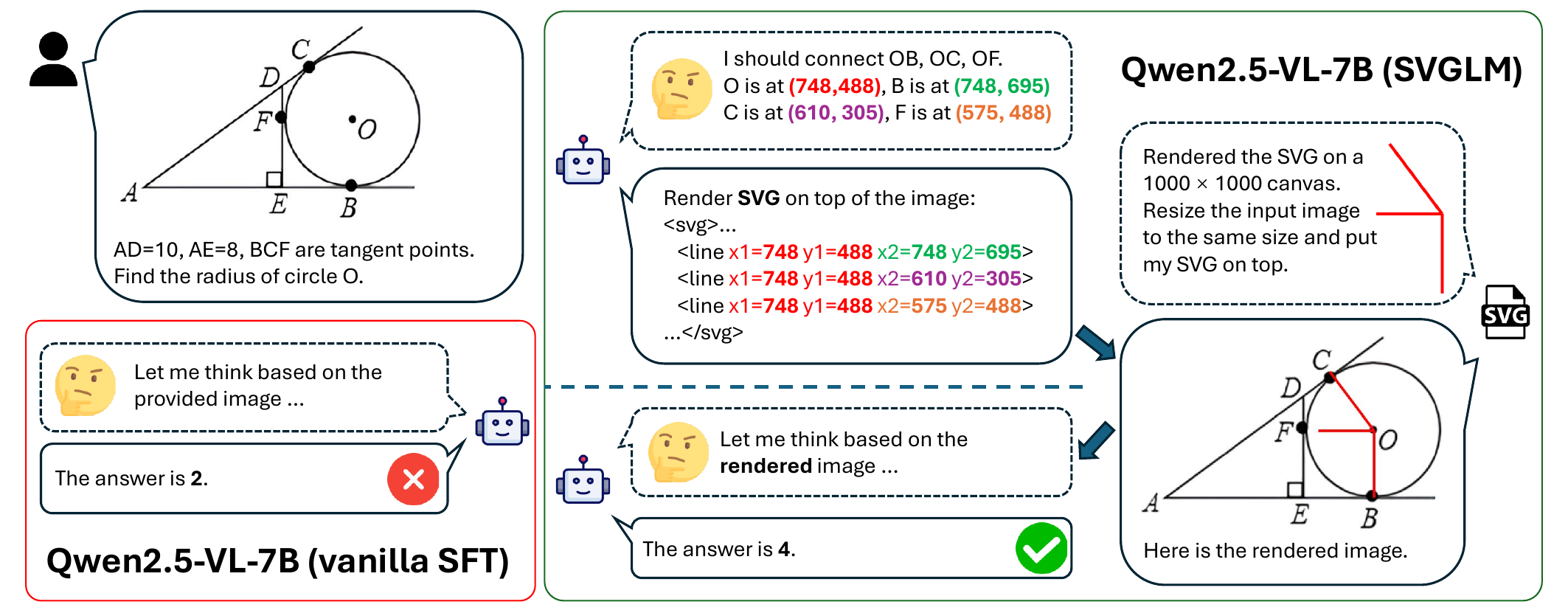}
  \caption{Comparison between standard multi-modal chain-of-thought and SVG-enhanced thinking. Vanilla SFT and SVGLM yield different conversation patterns and results. SVGLM offers a more robust thinking with images pipeline with better performance on reasoning tasks. }
  \label{fig:teaser}
\end{figure*}

\section{Introduction}

Vision-language models \citep[VLMs;][\textit{inter alia}]{radford2021learning,alayrac2022flamingo,team2023gemini,gpt4o,wang2024qwen2} have demonstrated promising performance in a range of downstream tasks~\citep{yue2024mmmu}, serving as an important building block of physical intelligence~\citep{huang2022inner,zitkovich2023rt,nasiriany2024pivot}.

However, two limitations remain salient for digital intelligence, where the visual input is often a graphical artifact rather than a natural photo.
First, current VLMs still struggle with infographics such as mathematical diagrams, charts, and structured visual documents~\citep{liu2023matcha,tang2025chartmuseum,shen2025recode}. 
These inputs compress symbolic structure, quantitative relations, and latent constraints into a small number of marks and coordinates.
In many cases, the model must understand the graphical structures behind the figure (e.g., axes, scales, layout rules) before reliable reasoning is possible.
Second, even when VLMs produce long and fluent rationales, the intermediate steps are frequently weakly grounded in the visual evidence~\citep{yao2025reasoning,liu2025more,zhang2025deepsketcher}. 
As a result, their conclusions can be dominated by linguistic priors and pattern completion rather than verified visual computation. 
This becomes acute in tasks where the missing ingredient is an explicit visual operation, e.g., adding an auxiliary line in geometry, where text-only reasoning is insufficient without an externalizable, checkable visual substrate.

Instead of relying on human-provided prompts, \citet{thinkwzimg}'s \textit{Thinking with Images} paradigm turns the image into an active workspace, where intermediate visual states are created and manipulated to support reasoning. 
Recent frameworks~\citep{hu2024visual,duan2025codeplot,qiao2025vthinker} operationalize this idea by treating visual manipulations as part of the reasoning process, e.g., generating auxiliary lines, marks, or plots as explicit intermediate steps. 
While some works aim to internalize such manipulations in latent space~\citep{zhang2025latent,zhang2025deepsketcher,wang2025monetreasoninglatentvisual}, explicit visual rationales remain attractive for interpretability and verifiability, especially in geometric and mathematical settings where step-by-step correction matters.

A practical question is the representation used to externalize these visual steps. 
Many existing systems either (i) generate raster or latent images, or (ii) generate imperative Python code that renders and edits figures. 
The former can be token-inefficient when interleaving high-resolution visual artifacts over long horizons, and the latter tightly couples reasoning to specific software stacks and runtime environments, which can be brittle under version drift and dependency changes.
For instance, VIGA and RECODE rely on Python programs executed in graphics engines or plotting tool chains for iterative render-and-revise loops~\citep{yin2026vision,shen2025recode}; CodePlot-CoT and DeepSketcher further train models to emit such code-based visual steps for geometry-centric reasoning~\citep{duan2025codeplot,zhang2025deepsketcher}. 
In contrast, renderable programs like Scalable Vector Graphics (SVG) offer a declarative, hierarchical, and resolution-independent interface: models can operate directly on compositional primitives (paths, shapes, groups, text) via structured text, making each visual modification both grounded and editable~\citep{yang2025omnisvg,wu2025chat2svg,rodriguez2025starvector}. 
This makes SVG a natural medium for visual reasoning as executable rationale: editable by construction, verifiable by rendering, and expressive enough to encode geometric structure without committing to heavyweight generation or tooling.

In this work, we present SVGLM, a SVG-enhanced reasoning paradigm to facilitate multimodal reasoning of VLMs, a paradigm to incorporate SVG as part of the thinking process into current vision language models (VLMs). Our contribution is summarized as follows:

\begin{compactitem}
\item We present a high-quality, LLM-verified dataset of SVG-as-image-editing samples with 8K samples.
\item We propose the first scheme to enhance multi-modal thinking with images rendered from vector programs.
\item We show that proper SVG integration achieves superior performance on trending multi-modal agents in solving downstream tasks.
\end{compactitem}

\section{Related Work}
\label{sec:background}

\subsection{Visual Representation by Graphics Programs}

A growing body of research has moved towards representing visual content through the lens of structured, editable graphics programs. Recent frameworks like VIGA~\citep{yin2026vision} and RECODE~\citep{shen2025recode} reconstruct images as executable programs, enabling a "write-run-render-compare-revise" loop to ensure geometric consistency. 
While VIGA leverages the deterministic nature of 3D engines such as Blender to bridge the semantic gap, RECODE focuses on decomposing visual content into 2D programmatic abstractions. 
Complementing these programmatic approaches, recent works~\citep{yang2025omnisvg,wu2025chat2svg,rodriguez2025starvector} leverage Scalable Vector Graphics (SVG) for visual content generation. 
Unlike rasterized pixels or the imperative code often used in general program synthesis, SVG provides a declarative, hierarchical, and resolution-independent representation.
This nature renders SVG an ideal medium for visual reasoning, as it allows models to manipulate high-level primitives (e.g., paths, shapes) directly through structured text, ensuring that every visual modification is inherently grounded in a verifiable and editable format.

\subsection{Visual Prompts and Thinking with Images}

VLMs exhibit visually grounded understanding of user-provided visual cues, motivating \textit{visual prompting}~\citep{yang2023set,yang2023dawn,feng2025visually}. 
Early work relied on tuning-based approaches~\citep{bahng2022exploring,yao2024cpt}, while later studies showed that VLMs can follow such cues zero-shot, e.g., overlaid marks and text~\citep{shtedritski2023does,yang2023dawn,li2023vrptest,yang2023set}.
A series of training-free methods have been proposed~\citep{lei2024scaffolding,yang2024fine,wan2024contrastive}, and recent work further strengthens visual prompting via visual instruction tuning~\citep{cai2023making} or explicit pointer tokens~\citep{lai2024lisa,zhang2024groundhog}. 
Instead of relying on human-provided prompts, \citet{thinkwzimg}'s \textit{Thinking with Images} paradigm transforms it into an active workspace where visual elements are dynamically manipulated to aid reasoning. 
Recent frameworks such as Visual Sketchpad~\citep{hu2024visual}, CodePlot-CoT~\citep{duan2025codeplot}, and V-Thinker~\citep{qiao2025vthinker} formalize visual prompts as a part of Chain-of-Thought reasoning. 
In these systems, models explicitly generate auxiliary lines, marks, or plots to externalize intermediate reasoning steps. 
While some recent works attempt to internalize this process into latent space~\citep{zhang2025latent,zhang2025deepsketcher,wang2025monetreasoninglatentvisual}, explicit Visual CoT offers distinct advantages in interpretability and precision, particularly for geometric and mathematical tasks. 
By externalizing the rationale onto a visual canvas, explicit methods allow for step-by-step verification and correction, a property essential for rigorous reasoning.

\subsection{Visual Reasoning with Code}

Visual reasoning via code generation leverages the modularity and precise logic of programming to address the compositional limitations inherent in end-to-end vision models. Early approaches~\citep{suris2023vipergpt, hu2024visual} facilitated this by composing vision models through generated Python subroutines. However, these methods are often constrained by a static, predefined library of vision modules, limiting their adaptability to novel tasks. More recently, agentic frameworks like PyVision~\citep{zhao2025pyvisionagenticvisiondynamic} and OpenThinkImg~\citep{su2025openthinkimglearningthinkimages} extend this by enabling MLLMs to synthesize bespoke Python tools for specific queries, grounding reasoning in executable code to handle complex geometric and document-based tasks. However, such general-purpose imperative code often abstracts geometric details behind procedural logic (e.g., hiding coordinates within variables or loops), creating a disconnect between the symbolic code and the actual spatial layout. This contrasts with SVG, which enforces explicit definitions of geometric attributes directly in the text, thereby intrinsically aligning the reasoning process with the visual result.
\section{Method}
\begin{figure*}
  \centering
  \includegraphics[width=\linewidth]{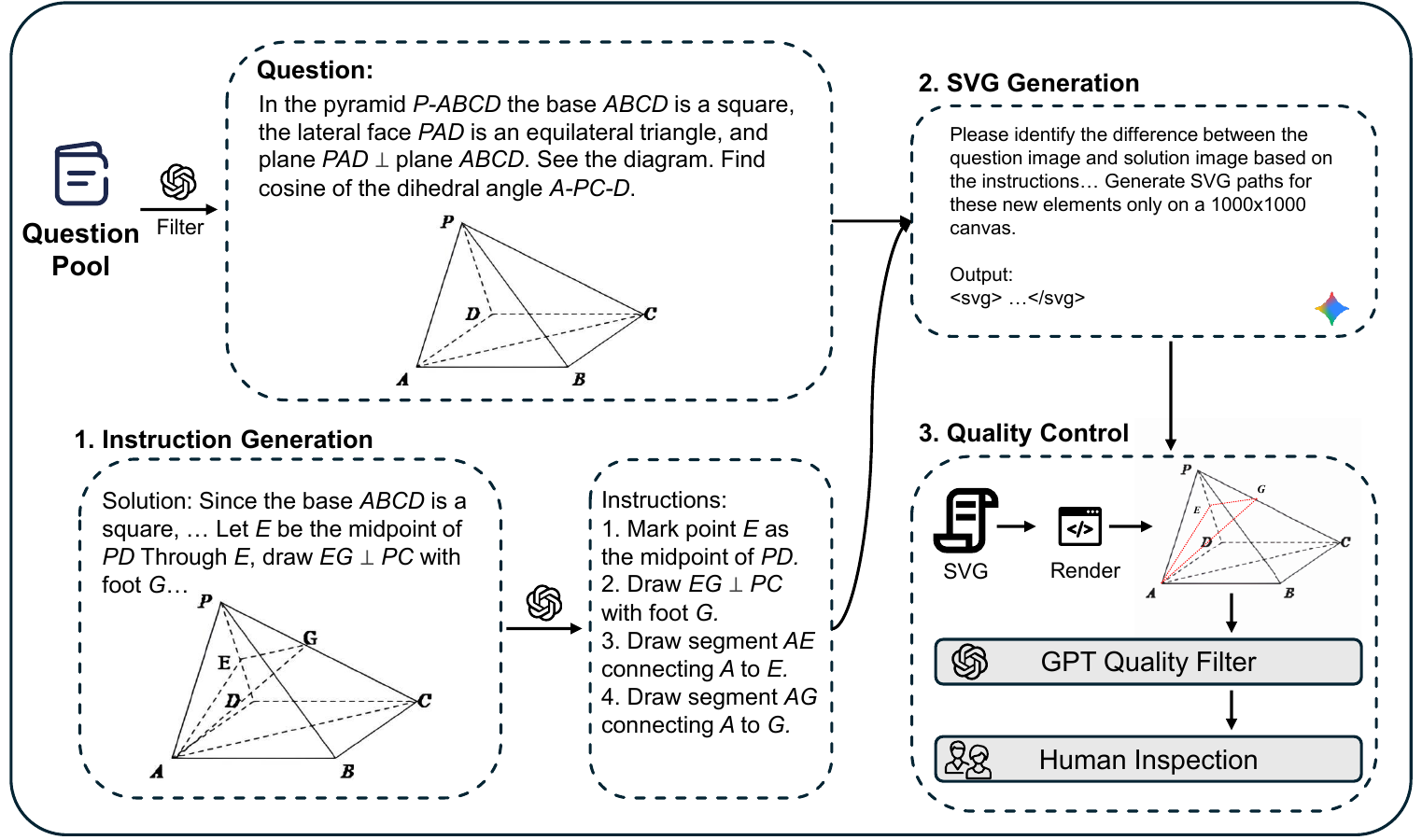}
  \caption{Our data curation pipeline. We use Gemini-3-Pro to identify the difference between source and target images and generate SVGs. Both GPT filtering and human verification are conducted to ensure the quality of collected samples.}
  \label{fig:data_gen_pipeline}
\end{figure*}

\subsection{Scalable Vector Graphics as Programs}

Scalable Vector Graphics (SVG)~\citep{w3c_svg11} is an XML-based vector graphics format used widely in icons, webpages, etc. It encodes scalable shapes, figures and texts into XML objects, where each entity is either a special object such as text or an embedded image, or a generic ``path'' formed by specifying the starting point and moving patterns. Tools like VTracer~\citep{vtracer} can decompose and encode any pixel-based image into SVG programs with arbitrary granularity, showing that SVG has ample visual representation power. Commercial language models like Gemini 3~\citep{deepmind2025gemini3} have shown strong abilities in understanding and generating SVGs consisting of simple shapes. These cases prove SVG as a bridge connecting XML-like text and free-form images.

In our work, we utilize SVG as a tool to effectively quantize and visualize the multi-modal thinking process. We emphasize vital properties of the connection between SVG and its rendered image: \textbf{grounded}, as entities like lines and rectangles in SVG will be clearly referenced in the resulting image; \textbf{editable}, since human and language model agents can easily manipulate SVG programs from XML-tree level without paying extra attention to grammatical or algorithmic failures; \textbf{semantically meaningful}, rising from the explicit coordinate representations of each entity in the XML codes in SVG. These properties are crucial in our application where multi-modal agents generate SVG code as part of the thinking process, while the deterministically rendered image from the given SVG is sent back to the agent, consistent with the standard multi-modal conversation pipeline.

\subsection{Thinking with Renderable Programs}

Modern vision-language models excel at understanding semantic visual information, but are not naturally equipped with multi-modal generation abilities. This limits their performance in tasks that require active visual editing and thinking based on the modified images. Omni models like Bagel~\citep{deng2025bagel} support multi-modal generation through unified pre-training, but lack the power to accurately reflect the thinking process in open-domain image editing. In our work, SVG is used as the medium bridging text-based thinking, which vision-language models are good at, and creating intermediate images that help chain-of-thought thinking. The structure of SVG programs allows surprisingly simple descriptions of standard figures on top of the canvas image, making it suitable for downstream tasks such as plotting and captioning crucial entities in complex visual understanding, or drawing auxiliary lines and shapes within the digital domain.

We introduce Scalable-Vector-Graphics-enhanced Language Models (SVGLM) to boost vision language models' multi-step reasoning abilities. We adopt the standard multi-turn tool-calling scheme, but allow normal text generation before the tool call. More precisely, the model can evoke a tool-call by wrapping arguments inside `<tool\_call>' and `<$\backslash$ tool\_call>' tokens any time during its generation process. The calling arguments include XML-encoded SVG and a specification of the image to be used as canvas, which can be either the question image or blank. Figure \ref{fig:teaser} shows a simplified interaction process, where the agent chooses to either render a new SVG or answer the original question in each turn of the conversation.

Our SVGLM pipeline can work seamlessly with any open-source trainable multi-modal agent. To empower any agent with renderable programs, we construct an 8K dataset of high-quality SVG tool-calling samples, which will be organized in conversation format consistent with the interactive reasoning pipeline mentioned above. A standard supervised fine-tuning is then conducted on each model to learn the structure and semantic relationship of SVGs. More detailed data collection and training are described in Section \ref{sec:data} and Section \ref{sec:train}, respectively.

\subsection{Data Curation}
\label{sec:data}
To equip VLMs with SVG reasoning ability, we curated a dataset consisting of 8,000 high-quality SVGs for drawing auxiliary lines in geometric problems. We use original images from MathCanvas-Instruct~\citep{shi2025mathcanvasintrinsicvisualchainofthought}, a large collection of multi-modal QA pairs where a ``solution image'' is crucial to answer the question. We select the \textit{plane geometry} and \textit{solid geometry} splits, as these categories have more samples that necessitates canvas-based geometric constructions instead of drawing from scratch.

As shown in Figure \ref{fig:data_gen_pipeline}, our SVG data collection pipeline proceeded in three stages:
\begin{compactitem}
    \item First, we employ GPT-5 to filter the source dataset, leaving problems where the solution image can be constructed from the question image by only adding captions and drawing new figures.
    \item Second, Gemini-3.0-Thinking is used to construct precise SVG annotations corresponding to intermediate visual steps. We let Gemini identify the difference between the question image and solution image, while accurately plotting each keypoint's location on a $1000\times 1000$ canvas. We choose Gemini-3.0-Thinking over GPT-5 as Gemini demonstrates stronger abilities in accurately deciding keypoint coordinates.
    \item Finally, we render generated SVGs onto original question images, and utilize GPT-5 as a visual verifier to assess the quality of the synthesis. Only samples with verified visual accuracy were retained.
\end{compactitem}

\begin{table}[h]
\centering
\small
\begin{threeparttable}
    \begin{tabular}{c c c c}
    \toprule
    \textbf{Perfect Reconstruction} & \textbf{Slight Translation} & \textbf{Major Translation} & \textbf{Structural Error} \\
    \midrule
    79.5\% & 10\% & 9.5\% & 1\% \\
    \bottomrule
    \end{tabular}
    \vspace{2pt}
    \caption{Human evaluation results in percentage.}
    \label{human}
\end{threeparttable}
\vspace{3pt}
\end{table}

To verify the quality of our collected samples, we conduct human evaluations on a random subset of our collected samples, asking if the constructed SVG reflects the shift between the question image and the solution image, as well as the accuracy of key points used by the SVG. We classify each sample into one of: perfect reconstruction, slight translational error, major translational error, and structural error, since translational discrepancy is not catastrophic in SVG generation. Evaluation percentages are indicated in Table \ref{human}, showing only $1\%$ of all evaluated samples have structural errors, thus confirming our dataset's quality.

For the multi-modal models' usage, we generate proper reasoning and tool calling with GPT-5, which is given a complete description and solution of each question, as well as the curated SVG and rendered image. We construct three different levels of reasoning density: no reasoning, concise reasoning and complete reasoning to suit different abilities of the target models. All generated messages are properly tokenized and organized to form our $3$-level SVG-enhanced conversation dataset.

\section{Experiments}

\begin{table*}
\centering
\small
\begin{threeparttable}
    \begin{tabular}{l c S S S S}
    \toprule
    \textbf{Model} & \textbf{Medium} & {\textbf{Plane Geometry}} & {\textbf{Solid Geometry}} & {\textbf{Weighted}} \\
    \midrule
    GPT-4o & - & 18.7 & 20.3 & 19.2 \\
    V-Thinker & Code & 19.0 & 23.8 & 20.5 \\
    GPT-4o + Qwen-Image-Edit & Image & 19.0 & 20.8 & 19.6 \\
    \midrule
    LLaVa-Next-Mistral-7B & - & 11.6 & 21.4 & 14.6 \\
    LLaVa-Next-Mistral-7B (SFT) & - & 9.8 & 20.7 & 13.2 \\
    LLaVa-Next-Mistral-7B (SVGLM) & SVG & \bfseries 20.3 & \bfseries 29.3 & \bfseries 23.1 \\
    \midrule
    Qwen2.5-VL-7B & - & 19.1 & 20.5 & 19.5 \\
    Qwen2.5-VL-7B (SFT) & - & 20.7 & 24.2 & 21.8 \\
    Qwen2.5-VL-7B (SVGLM) & SVG & \bfseries 30.0 & \bfseries 33.0 & \bfseries 30.9 \\
    \midrule
    InternVL3-8B & - & 18.9 & 20.7 & 19.5 \\
    InternVL3-8B (SFT) & - & 22.5 & 24.1 & 23.0 \\
    InternVL3-8B (SVGLM) & SVG & \bfseries 28.7 & \bfseries 32.4 & \bfseries 29.8 \\
    \bottomrule
    \end{tabular}
    \vspace{1.5pt}
    \caption{Comparative evaluation of multi-modal agents on MathCanvas-Bench. Scores represent weighted accuracy. Best results within each model family are highlighted.}
    \label{tab:geometry_bench_updated}
\end{threeparttable}
\end{table*}

\subsection{Dataset and Metric}

We use MathCanvas-Bench~\citep{shi2025mathcanvasintrinsicvisualchainofthought} as the benchmark to evaluate reasoning ability that benefits from drawing auxiliary lines. MathCanvas-Bench consists of 3000 question-answer pairs over 8 math categories that require multi-step visual thinking, each of which consists of one to three sub-questions. We select Plane Geometry (1092 samples) and Solid Geometry (486 samples) from the benchmark's splits for evaluation, in accordance with our curated dataset.

To reflect rendering images on a canvas, we filter only samples from the above benchmark with at least one image in question, leaving a total of 1244 (78.8\%) QA pairs. We report accuracy over both categories, as well as a weighted overall score. Following MathCanvas-Bench, we calculate accuracy with weighted scoring that rewards correct answers of later sub-questions more heavily, where each sub-question's weight is $30\%$ larger than the previous sub-question.

\subsection{Evaluation setting}
\label{sec:train}
We select $3$ open-source models of roughly the same size: LLaVa-Next-Mistral-7B~\citep{liu2024llavanext}, Qwen2.5-VL-7B-Instruct~\citep{Qwen2.5-VL} and InternVL-3-8B~\citep{zhu2025internvl3}. Despite similar scale and architectures, these models vary in zero-shot abilities like multi-step mathematical reasoning and SVG generation. All selected models are tested against three settings:
\begin{compactitem}
    \item \textbf{Zero-shot}: The models are given the original question and image without training. They then answer each question after Chain-of-Thought thinking~\citep{wei2022chain}.
    \item \textbf{SFT}: We collect questions from MathCanvas-Instruct~\citep{shi2025mathcanvasintrinsicvisualchainofthought} as user messages with the same number of samples as our SVG-enhanced conversation dataset. GPT-5-generated chain-of-thought reasoning is conducted with rejection-sampling using questions from MathCanvas-Instruct, after which supervised-finetuning is applied on each model.
    \item \textbf{SVGLM}: Selected models are fine-tuned on our SVG-enhanced conversation dataset, respectively. All models are evaluated in the interactive tool-calling paradigm on MathCanvas-Bench.
\end{compactitem}

Each separate SFT experiment is conducted with 3 epochs, learning rate $10^{-5}$ and batch size $8$ on 8 NVIDIA-H100 GPUs. Running time of different models varies slightly, but all experiments can finish after $2$ hours. Detailed training setups are described in the appendix.

We also select several baseline methods for comparison as follows:

\textbf{V-Thinker}~\citep{qiao2025vthinker} uses python coding as tool-based visual chain-of-thought. We use the officially released pre-trained model and pipeline of V-Thinker on our dataset.

\textbf{Qwen-Image-Edit}~\citep{wu2025qwenimagetechnicalreport} is an open-source image editing model. We use GPT-4o as the backbone to generate clear image editing instructions such as "connect A and B in this image", whose result is then fed into Qwen-Image-Edit-2511 along with the starting canvas to synthesize an image. This is analogous to the \textbf{SVGLM} pipeline with GPT-4o's instructions replacing SVG generation and Qwen-Image-Edit replacing the SVG renderer. 

\textbf{Commercial models} We evaluate GPT-4o under the zero-shot setting to illustrate the difficulty of the MathCanvas benchmark.

\subsection{Generated SVGs}

\begin{figure*}
    \centering
    \includegraphics[width=0.9\textwidth]{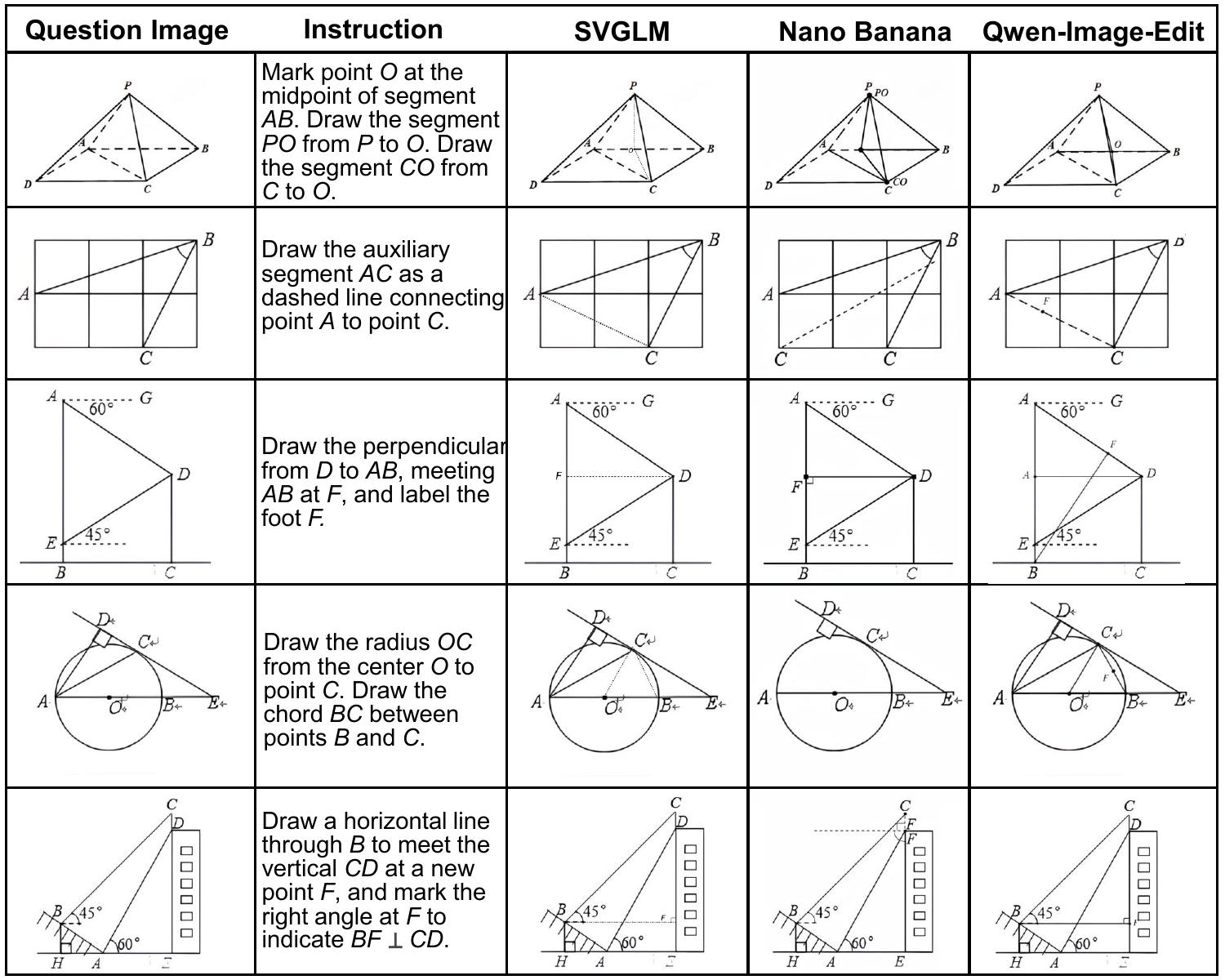}
    \caption{Instruction and SVG generated by SVGLM, as well as Qwen-Image-Edit and Nano Banana on the exact same prompt}
    \label{fig:generated}
\end{figure*}

\label{sec:qualitative}
We show qualitative results of reconstructed SVGs from our trained models in Figure \ref{fig:generated}, along with image editing results from both Qwen-Image-Edit~\citep{wu2025qwenimagetechnicalreport} and Nano-Banana ~\citep{google2026gemini3pro} with the same instruction. We can see that large image editing models such as Qwen-Image-Edit or Nano-Banana suffer from severe hallucination and domain misalignment, making it difficult to follow the text instructions. Moreover, fine-tuning on diffusion-based image editing models requires significantly more data compared to standard VLM fine-tuning, which blocks the path of adapting them to efficient thinking with images.

Figure \ref{fig:generated} also verifies the base model's ability to adjust and construct useful instructions and high-quality SVGs through fine-tuning. We find various types of SVG drawing, including connecting two points, spanning a parallel/perpendicular line from a point, marking an intersection with a letter, and indexing important angles. The diversity of SVG behavior demonstrates the potential of VLMs as primitive vector graphics generators.

\subsection{Results}

\begin{table*}
\centering
\small
\begin{threeparttable}
    \begin{tabular}{c S S S S}
    \toprule
    {\textbf{Model}} & {\textbf{Plane Geometry}} & {\textbf{Solid Geometry}} & {\textbf{Weighted}} \\
    \midrule
    Qwen2.5-VL-7B w/o solution image & 19.1 & 20.5 & 18.8 \\
    Qwen2.5-VL-7B w solution image & 33.9 & 34.7 & 34.0 \\
    \bottomrule
    \end{tabular}
    \vspace{2pt}
    \caption{Ablation study of the presence of solution images. All accuracies are calculated the same way as Table \ref{tab:geometry_bench_updated}.}
    \label{tab:ablation3}
\end{threeparttable}
\end{table*}

A summary of our experiments is shown in \textbf{Table \ref{tab:geometry_bench_updated}}. From the table, it is evident that SVGLM achieves better results on our benchmark with all three base models. While zero-shot and direct SFT on base models yield comparative results to GPT-4o, SVGLM surpasses both GPT-4o and open-source thinking-with-images baselines by a large margin. We find the following points worth noting:
\begin{compactitem}
    \item GPT-4o with Qwen-Image-Edit as the reasoning medium does not improve over using GPT-4o only. As shown in Section \ref{sec:qualitative}, this is due to the hallucination and low-quality outputs from Qwen-Image-Edit. 
    \item We notice a performance decrease on several entries between zero-shot and direct SFT. Admittedly, different models demonstrate different sensitivity levels to prompt structures and chain-of-thoughts in training, but only to a marginal error. Further discussion is done in Section \ref{sec:ablation}. Most experiments see a reasonable performance improvement on direct SFT, considering the reasoning abilities already present in selected models.
\end{compactitem}

Overall, we see SVGLM with a significant boost of $8\%$ over zero-shot and $6.5\%$ over direct SFT. It also beats GPT-4o and the two think-with-images baselines.

\subsection{Generalization to grounded reasoning}

Further experiments are conducted to showcase SVG's power in grounded reasoning, where the answer must ground on specific pixels or regions in the image. We select 30,000 random samples from Visual Genome~\citep{krishna2016visualgenomeconnectinglanguage}, consisting of visual QA pairs concerning specific objects in the image. The bounding box in each sample is converted so that the rendered SVG contains a visible box on top of the original image. This selected dataset is used to reproduce the whole training process of SVGLM.

We then test our trained models on POPE~\citep{li2023evaluatingobjecthallucinationlarge}, a dataset to evaluate vision-language models' hallucination. POPE's samples are hard to answer if not concretely grounded on specific objects in the image, making them a suitable candidate to solve with our canvas-based method. Results are in Table \ref{tab:grounded}, with all three models seeing improvement from vanilla SFT. Although grounded reasoning's used SVGs are simple and equivalent to explicitly drawing boxes, model can benefit from the generalizability of SVGs to adapt to broader tasks and shapes.

\begin{table*}[!h]
\centering
\small
\begin{threeparttable}
    \begin{tabular}{l c c c}
    \toprule
    \textbf{Model} & \textbf{Zero-shot} & \textbf{SFT} & \textbf{SVGLM} \\
    \midrule
    LLaVa-Next-Mistral-7B & 47.4 & 61.2 & \textbf{69.4} \\
    Qwen2.5-VL-7B & 74.4 & 81.9 & \textbf{83.8} \\
    InternVL3-8B & 81.6 & 82.3 & \textbf{84.8} \\
    \bottomrule
    \end{tabular}
    \vspace{1.5pt}
    \caption{Evaluation results of grounded reasoning with POPE benchmark}
    \label{tab:grounded}
\end{threeparttable}
\end{table*}

\subsection{Ablation studies}
\label{sec:ablation}
We conduct several ablation studies to strengthen the rationale of the SVGLM pipeline.

\textbf{Model choices} Our three models are chosen as representatives of different model families and training strategies. We notice serious training data contamination starting from Qwen3-VL-8B~\citep{bai2025qwen3vltechnicalreport} and InternVL-3.5-8B~\citep{wang2025internvl35advancingopensourcemultimodal}. However, we found that: zero-shot inference of Qwen3-VL-8B achieves 30.8\% on the benchmark, while pure SFT on the same set of prompts scored 22.8\% which is lower than zero-shot; on one category of the dataset, Trigonometry, zero-shot model has 49.2\% accuracy, significantly higher than GPT-4o and SFT results, which yields around 20\%. This performance degradation suggests potential data contamination with the testing data. Therefore, we choose Qwen2.5-VL-7B and InternVL-3-8B as they are among the most powerful open-sourced vision language models by the time the benchmark was released at the 7B scale.

\textbf{Is auxiliary images necessary?} This question arises from our benchmark naturally, since the difficulty of samples with ground truth solution images is an apparent upper bound of how SVGs can perform given infinite budget of text-based inference. To quantitatively analyze this, we first generate ground truth image editing instructions using the question image and solution image pairs; then, we test zero-shot performances of Qwen2.5-VL-7B on MathCanvas-Bench, given both images alongside the generated text instruction. Table \ref{tab:ablation3} illustrates the impact of solution images' presence on zero-shot Qwen2.5-VL-7B model. It is evident that the questions do appear much simpler to multi-modal agents when come with the solution image.

\textbf{Ablation of Framework Components.} We verify the contributions of all components within our SVGLM framework by isolating the effects of SVG generation and rendering. To maintain consistency with the supervised fine-tuning (SFT) baselines, we utilize Gemini exclusively to generate the SVGs, while the Chain-of-Thought (CoT) reasoning is generated by GPT-5 across the same set of prompts. However, enforcing perfectly identical prompts across all configurations is not strictly plausible; the full CoT unavoidably incorporates visual information from the rendered SVG. Consequently, omitting the SVG removes crucial context that standard SFT relies on.

To decouple these mechanisms and strengthen our argument, we evaluate models under three distinct settings: the full SVGLM framework, SVGLM without the rendered SVG feedback (forcing the model to output the solution based only on the generated SVG code), and SVGLM without any SVG generation or rendering. As Table \ref{tab:ablation1} indicates, thinking with SVGs (w/o SVG render) already shows improvements over settings lacking both generation and rendering. Incorporating the rendered images back into the context pushes the performance even further. For instance, removing both SVG generation and rendering on Qwen2.5-VL significantly degrades performance, demonstrating that both the generation and rendering phases are crucial to our method. While LLaVa-Next sees a comparatively smaller boost due to its limited capacity in multi-image understanding, all models achieve their highest performance with the full SVGLM equipped. This solidifies our argument that SVGLM's dual mechanism of creation and visual reflection fundamentally charges VLM reasoning abilities.

\begin{table}[!htbp]
\centering
\small
\begin{threeparttable}
    \begin{tabular*}{\linewidth}{@{\extracolsep{\fill}} l S S S S @{}}
    \toprule
    \textbf{Model} & {\textbf{SFT}} & {\textbf{SVGLM}} & {\textbf{w/o SVG render}} & {\textbf{w/o SVG gen. \& render}} \\
    \midrule
    LLaVa-Next   & 11.6 & \textbf{20.0} & 19.5 & 10.7 \\
    Qwen2.5-VL   & 20.3 & \textbf{26.8} & 23.8 & 20.2 \\
    InternVL3    & 21.7 & \textbf{26.2} & 23.8 & 21.4 \\
    \bottomrule
    \end{tabular*}
    \vspace{2pt}
    \caption{Ablation study verifying all framework components. All figures represent the weighted score on MathCanvas-Bench. \textit{w/o SVG render} forces the model to output the answer after writing the SVG code, without the rendered image as feedback. \textit{w/o SVG gen. \& render} represents the baseline without any SVG assistance.}
    \label{tab:ablation1}
\end{threeparttable}
\end{table}

\textbf{Effect of different prompts} We assess the outcome of different reasoning densities in our curated dataset, namely: \textit{No reasoning} only keeps the instruction for the SVG as well as steps to solve the original problem; \textit{Concise} gives moderate steps of reasoning; \textit{Complete} is a detailed, step-by-step reasoning on both SVG generation and question answering. Table \ref{tab:ablation2} shows the experimental results on the same benchmark. clearly indicating the impact of prompt patterns in SFT training. Presumably, models with stronger reasoning abilities will prefer shorter answers as SFT, as longer prompts are likely inconsistent with the models' internal knowledge, thus more resources are used in shifting the distribution of text generation instead of focusing on answering correctly. However, we mark that whichever set of reasoning densities produce a significant increase in weighted accuracies of our benchmark.

\begin{table}[htbp]
\centering
\small
\begin{threeparttable}
    \begin{tabular}{l S S S}
    \toprule
    \textbf{Model} & {\textbf{No Reasoning}} & {\textbf{Concise}} & {\textbf{Complete}} \\
    \midrule
    LLaVa-Next  & 18.0 & \textbf{20.0} & 19.0 \\
    Qwen2.5-VL   & \textbf{26.8} & 23.9 & 23.4 \\
    InternVL3    & \textbf{26.2} & 24.8 & 22.4 \\
    \bottomrule
    \end{tabular}
    \vspace{2pt}
    \caption{Ablation study of reasoning density. All figures are the weighted score of MathCanvas-Bench. }
    \label{tab:ablation2}
\end{threeparttable}
\end{table}

\section{Conclusion}

In this paper, we present SVGLM, a paradigm to boost multi-modal agents with the ability to think with renderable programs, as well as a large fine-tuning dataset of 8K high-quality question-SVG pairs. We show that with proper tool-calling integration, multi-modal agents with weak reasoning abilities can learn to think with vector graphics as renderable programs, exhibiting better performance than other thinking with images baselines as well as standard fine-tuned VLMs. On the other hand, SVGLM is suitable for Document QA and grounded reasoning tasks as well, where SVG can be used to draw charts and plot important entities. The success in both math and spatial reasoning demonstrate the potential of SVGLM as an easy and elegant solution to boosting smaller models' reasoning abilities.

\textbf{Limitations} Due to lack of compute resources and paid API requests, we're not able to verify our results in larger models and larger SVG datasets. We will try our best to scale up SVGLM, as well as applying inference-time scaling methods in subsequent works.



\bibliography{refer}
\bibliographystyle{tmlr}

\appendix

\section{Dataset}
\subsection{Prompt design for SVG data generation}
\label{app:prompt for svg}
\begin{figure}[ht]
    \centering
    \begin{promptbox}
        Your task is to identify the difference between the Question image and solution image based on the instructions. You should generate a transparent overlay in SVG format. Do NOT redraw existing lines from the Question Image. Output ONLY the new auxiliary constructions. Coordinates must align perfectly with the Question Image boundaries.

        \vspace{10pt}
        \noindent 1. Inputs: \\
        - Question Image: The base diagram. \\
        - Solution Image: The reference solution diagram. \\
        - Instructions: ``\textcolor{cyan}{\{instruction\}}''

        \vspace{10pt}
        \noindent 2. Thinking Process \\
        - Detect: Identify existing points (anchors) in the Original Image. \\
        - Map: Convert anchor positions to 1000x1000 coordinates. \\
        - Construct: Calculate coordinates for new lines based on the instruction: ``\textcolor{cyan}{\{instruction\}}''.

        \vspace{10pt}
        Output the SVG and step-by-step reasoning in the required JSON structure.
    \end{promptbox}
    \caption{Template used for SVG generation task. The text highlighted in \textcolor{cyan}{cyan} indicates variables to be replaced.}
    \label{fig:prompt-template}
\end{figure}

\subsection{Prompt design for quality assessment}
\begin{figure}[ht]
    \centering
    \begin{promptbox}
        You are given three images: \\
        1. Input Image: The original question image from a geometry problem \\
        2. Target Image: The expected solution/answer image \\
        3. Output Image: The generated/reconstructed SVG image rendered as PNG \\
        
        Please analyze these images and determine if the output image is acceptable for use in educational geometry content.
        
        \vspace{10pt}
        Evaluation Criteria: \\
        - Visual Quality: Is the output image clear and well-rendered? \\
        - Correctness: Does the output match the key geometric elements in the target? \\
        - Completeness: Are all important geometric shapes, lines, and labels preserved? \\
        - Usefulness: Would this output be helpful for students learning geometry? \\
        
        \vspace{10pt}
        Respond with a JSON object containing: \\
        \{ \\
        \quad "accept": true/false, \\
        \quad "reason": "Brief explanation of your decision (2-3 sentences)", \\
        \quad "quality\_score": 1-10, \\
        \quad "geometric\_correctness": 1-10, \\
        \quad "visual\_clarity": 1-10 \\
        \} \\
        
        \vspace{10pt}
        Be strict in your evaluation - only accept outputs that accurately represent the geometric content.
    \end{promptbox}
    \caption{Template used for SVG quality assessment.}
    \label{fig:prompt-template2}
\end{figure}

\section{Training setup}

We adopted LLaMa-Factory~\citep{zheng2024llamafactory} as the training base for LLaVa-Next-Mistral-7B~\citep{liu2024llavanext}, Qwen2.5-VL-7B-Instruct~\citep{Qwen2.5-VL} and InternVL-3-8B~\citep{zhu2025internvl3}. Our system for training has Intel(R) Xeon(R) Platinum 8468 CPU and 8 Nvidia H100 GPUs. We use the full fine-tune setting that tunes all language model parameters while freezing all three VLMs' vision towers and multi-modal projectors. We use cosine learning rate scheduler with $10^{-5}$ learning rate. All models are trained for $3$ epochs under each setting, with per-device batch size $1$ (effectively batch size $8$).

\end{document}